\documentclass[runningheads]{llncs}
\usepackage{times}
\usepackage{latexsym}
\usepackage{hyperref}
\usepackage{booktabs}
\usepackage{amsmath}
\usepackage{graphicx}
\usepackage{mathptmx}
\usepackage{xcolor}
\usepackage{textcomp}
\usepackage{subfig}
\usepackage{soul,xcolor}
\usepackage[draft]{todonotes}
\usepackage{multirow}
\usepackage[bottom]{footmisc}
\usepackage{float}
\usepackage{placeins}
\usepackage{comment}

\usepackage{url}

\newcommand{\removecolors}{1}

\if \removecolors 1
\newcommand{\af}[1]{{#1}}
\newcommand{\jl}[1]{{#1}}
\newcommand{\pt}[1]{{#1}}

\newcommand{\nc}[1]{{#1}}
\newcommand{\dl}[1]{{#1}}
\else
\newcommand{\jl}[1]{\textcolor{red}{#1}}

\newcommand{\af}[1]{\textcolor{blue}{#1}}

\newcommand{\pt}[1]{\textcolor{red}{#1}}

\definecolor{deniscolor}{HTML}{2098A8}
\newcommand{\nc}[1]{\textcolor{orange}{#1}}

\newcommand{\dl}[1]{\textcolor{deniscolor}{#1}}

\fi

\newcommand{\datasparse}{limited-data}
\newcommand{\simplequestions}{\textsc{SimpleQuestions}}

\renewcommand{\vec}[1]{\mathbf{{#1}}}

\begin{document}

%\aclfinalcopy % Uncomment this line for the final submission
%\def\aclpaperid{***} %  Enter the acl Paper ID here

%\setlength\titlebox{5cm}
% You can expand the titlebox if you need extra space
% to show all the authors. Please do not make the titlebox
% smaller than 5cm (the original size); we will check this
% in the camera-ready version and ask you to change it back.

\title{Pretrained Transformers for Simple Question Answering over Knowledge Graphs}
\author{Denis Lukovnikov\inst{1} \and Asja Fischer\inst{2} \and Jens Lehmann\inst{1,3}}
\institute{
University of Bonn, Germany\\
\email{\{lukovnik,jens.lehmann\}@cs.uni-bonn.de}\\
\and
Ruhr University Bochum, Germany \\
\email{asja.fischer@rub.de} \\
\and 
Fraunhofer IAIS, Dresden, Germany\\
\email{jens.lehmann@iais.fraunhofer.de}
}

\authorrunning{Lukovnikov et al.}

\maketitle              % typeset 

%\title{Instructions for ACL 2018 Proceedings}

%\author{First Author \\
%  Affiliation / Address line 1 \\
%  Affiliation / Address line 2 \\
%  Affiliation / Address line 3 \\
%  {\tt email@domain} \\\And
%  Second Author \\
%  Affiliation / Address line 1 \\
%  Affiliation / Address line 2 \\
%  Affiliation / Address line 3 \\
%  {\tt email@domain} \\}

%\date{}

%\begin{document}

%\maketitle
\begin{abstract}
Answering simple questions over knowledge graphs is a well-studied problem in question answering.
Previous approaches for this task built on recurrent and convolutional neural network based architectures that use pretrained word embeddings.
It was recently shown that finetuning pretrained transformer networks (e.g. BERT) can outperform previous approaches on various natural language processing tasks.
In this work, we investigate how well BERT performs on \simplequestions{} and provide an evaluation of both BERT and BiLSTM-based models in \datasparse\  scenarios.
\end{abstract}

%%%%%%%%%%%%%%%%%%%%%%%%%%%%%%%%%%%%%%%%%%%%%%%%%%%%%%%%%%%%%%%%%%%%%%%%%%%%%%%%
\section{Introduction}
Question Answering (QA) over structured data aims to directly provide users with answers to their  questions (stated in natural language), computed from data contained in the underlying database or knowledge graph (KG).
To this end, a knowledge graph question answering (KGQA) system has to understand the intent of the given question, formulate a query, and retrieve the answer by querying the underlying knowledge base. 
The task of translating \dl{natural language (NL)} inputs to their logical forms (queries) is also known as semantic parsing.   
In this work, we focus on answering simple questions (requiring the retrieval of only a single fact) over KGs such as Freebase~\cite{freebase}.  
%To this end, \af{and therefore} use the \simplequestions\ dataset.\todo{may be remove this. There is a whole paragraph worth of content in the introduction later}

The availability of large quantities of high-quality data is essential for successfully training neural networks on any task. However, in many cases, such datasets can be difficult and costly to construct. Fortunately, the lack of data can be mitigated by relying on transfer learning from other tasks with more data. 
In transfer learning, (neural network) models are first trained on a different but related task, with the goal of capturing relevant knowledge in the pretrained model. Then, the pretrained model is finetuned on the target task, with the goal of reusing the knowledge captured in the pretraining phase to improve performance on the target task.
 
%%%%%%% ALTER VERSION START %%%%%%%%%%%%%%%
%\dlnote{ALTERNATIVE START}
Recently proposed transfer learning methods~\cite{ulmfit,openaigpt,openaigpt2,bert,elmo,mtdnn} show that significant improvement on downstream \dl{natural language processing (NLP)} tasks can be obtained by finetuning a neural network that has been trained for language modeling (LM) over a large corpus of text data without task-specific annotations.
% Benefit also is faster convergence and data sparse thing.
Models leveraging these techniques have also shown faster convergence and encouraging results in a few-shot or \datasparse\ settings~\cite{ulmfit}.
Owing to their benefit, the use of this family of techniques is an emerging research topic in the NLP community~\cite{mtdnn}.
% this family of techniques is an emerging research topic in the NLP community~\cite{bertmultitask}.
% Owing to their benefit, this family of techniques is an emerging research topic in the NLP community~\cite{bertmultitask}.
% While this family of techniques have 
However, it has received little attention in KGQA research \dl{so far}.

% One such \ptrm{state-of-the-art} pretrained neural network for NLP is BERT~\cite{bert}, which is a deep transformer~\cite{transformer} network trained with a masked language model\pt{ing} (MLM) task as well as a sentence pair classification task.
% \pt{\st{Transformers are a recently proposed architecture that relies on multi-head self-attention to encode natural language (NL) input. 
% Transformers are radically different from the common RNN and CNN architectures, as we will elaborate in Section~\ref{sec:transformers} and are argued to handle long-range dependencies in text better because of the use of attention rather than repeated state updates (RNN) or local linear transformations with global pooling operations (CNN).}}

%In this work, we investigate the use  the \af{performance} of BERT~\cite{bert}, a (bidirectional) transformer~\cite{transformer} pretrained on both a Masked LM task and a sentence pair classification task, \af{for KGQA}.

The main focus of our work is to investigate
%\todo{investigate what? Performance,robustness,effectiveness} 
transfer learning for question answering over knowledge graphs (KGQA) using models pretrained for language modeling.
% So far, transfer learning \pt{has} received little attention in KGQA research while it can significantly boost performance of KGQA systems and is an interesting emerging research topic.
For our investigation, we choose BERT~\cite{bert} as our pretrained model used for finetuning, and investigate transfer from BERT using the \simplequestions~\cite{bordes2015large} task.
\jl{BERT} is a deep transformer~\cite{transformer} network trained on
a masked language modeling (MLM) task as well as a \jl{subsequent} sentence pair classification task.
%To the best of our knowledge, transformer-based architectures for KGQA and transfer from LM-pretrained transformer networks for \simplequestions\ have not been investigated before.
We use \simplequestions\ because it is a very well-studied dataset that characterizes core challenges of KGQA, and is, to the best of our knowledge, the largest gold standard dataset for KGQA.
The large size of the dataset is particularly appealing for our study, because it allows us to investigate performance for a wider range of sizes of the data used for training.
Promising results with BERT for KGQA have been very recently reported on other KGQA datasets~\cite{maheshwari2019}. However, %existing work lacks a 
\af{we found a}
thorough investigation \af{of} the impact of data availability and an analysis of internal model behavior \af{to be missing, which} would help \af{to} better understand model behavior in applications \af{of} KGQA.

The contributions of this work are as follows:

%To resolve these shortcomings in the current state of research, we offer the following contributions in this work:
\begin{itemize}
\item We \dl{demonstrate for the first time the use of \af{a} pretrained transformer network (BERT) for simple \af{KGQA}.}%question answering over knowledge graphs.
%propose a novel neural network based approach for simple question answering over knowledge graphs, based on pretrained transformer networks (BERT). 
We also propose a simple change in our models that yields a significant improvement in entity span prediction compared to previous work.
\item We provide a thorough evaluation of pretrained transformers on \simplequestions\ for different amounts of data used in training
%under different data availability cases 
and compare with a strong baseline based on bidirectional Long-Short-Term-Memory~\cite{lstm} (BiLSTM). 
To the best of our knowledge, our work is the first to provide an analysis of performance degradation with reduced training data sizes for \simplequestions\ and KGQA in general.
\item We try to provide an understanding of the internal behavior of transformer-based models by analyzing the changes in internal attention behavior induced in the transformer during finetuning. %\dlnote{TODO: add explainable AI sentence}
\end{itemize}

%In this work, we provide (1) an evaluation of BERT-based entity detection and relation classification, compared to a strong \af{bidirectional Long-Short-Time-Memory (}BiLSTM) baseline, (2) a data ablation (???) study comparing our BERT-based models to the BiLSTM baseline models.
%To the best of our knowledge, \af{BERT has not been evaluated for \simplequestions\ or KGQA in general} yet,  \af{and we are the  first to provide an detailed} analysis of performance degradation with reduced training data size %has not been performed yet 
%for \simplequestions, with only few works reporting \af{a} similar analyses for other related datasets~\cite{TODO}.
%In addition, to the best of our knowledge, BERT has not been evaluated for \simplequestions\ or question answering over knowledge graphs in general.

We perform our study using the general framework used in recent works~\cite{buboqa,nearlysolved}, where simple question interpretation is decomposed into (1) entity span detection, (2) relation classification, and (3) a heuristic-based post-processing step to produce final predictions.
\dl{In this work, we particularly focus on the first two subtasks, providing detailed evaluation results and comparison with a baseline as well as \cite{buboqa}.
%since they form the core of the learning problem and their performance can be more easily studied in isolation, as it is not affected by heuristics and post-processing.
}

\section{Approach}
We follow the general approach outlined in BuboQA~\cite{buboqa}, which decomposes simple question interpretation into two separate learning problems: (1) entity span detection and (2) relation classification.
Recent works on \simplequestions\ show that this general approach, followed by a heuristic-based entity linking and evidence integration step can achieve state-of-the-art performance~\cite{buboqa,nearlysolved}, compared to earlier works~\cite{lukovnikov2017,dai2016,golub2016,yu2017improved} that investigated more complicated models.
%
%In our work, we focus on the two learning subtasks since they form the core of the learning problem and their performance can be more easily studied in isolation, as it is not affected by further heuristics and postprocessing.

In summary, our approach follows the following steps at test time:
\begin{enumerate}
    \item \jl{Entity Span Detection and Relation Prediction}: %\af{[remove:$P(s, r|q)$]}
    The fine-tuned BERT model is used to perform sequence tagging to both (1) identify the span $s$ of the question $q$ that mentions the entity (see Section~\ref{sec:approach:spans}) and (2) \af{predict} the relation $r$ used in $q$ (see Section~\ref{sec:approach:relations}). 
    In the example ``\textit{Where was Michael Crichton born?}'', $s$ would be the span ``\textit{Michael Crichton}''. 
    The tagger is trained using annotations automatically generated from the training data and entity labels in Freebase.
    \item Entity Candidate Generation: 
    %all entities are selected whose names \jl{are similar to} 
    %the predicted entity span $s$. 
    We retrieve entities whose labels are similar to the predicted entity span $s$ using an inverted index\footnote{The inverted index maps words to entities whose labels contain that word} and rank them first by string similarity (using \texttt{fuzzywuzzy}) and then by the number of outgoing relations.
    %See Section~\ref{sec:approach:final}.
    For our example, the resulting set of entity candidates will contain the true entity for Michael Crichton, the writer (corresponding to Freebase URI http://www.freebase.com/m/056wb). Note that the true entity does not necessarily rank highest \dl{after} \af{the} retrieval phase.
    \item \pt{Q}uery \pt{R}anking: 
    %\pt{\st{given the relations predicted in Step 1, the set of entities from Step 2 is ranked according the string similarity, best connected relation probability and in-degree of the entity. See Section~\ref{sec:approach:final}.}}
    \jl{Given} the relations predicted in Step 1, and the set of entities from Step 2, the entity-relation pairs are re-ranked as detailed in Section~\ref{sec:approach:final}.
    \dl{After ranking entity-relation pairs, we take the top-scoring pair, from which we can trivially generate a query to retrieve the answer from the KG.}
    %of to the co-occurence of the entities with relation $r_{max}$ in the KG. The top-ranked entity is used in the predicted query.
    %relation is predicted and given that relation, an entity is predicted such that the entity has the name $a$ and has the predicted relation.
\end{enumerate}

%Instead of using RNNs for both tasks, we instead employ BERT~\cite{bert}, a pretrained transformer~\cite{transformer}.

%Where
Whereas previous works experimented with \dl{recurrent and convolutional neural network (RNN resp. CNN)} architectures, we investigate an approach based on transformers. 
Several existing works train separate models for the two learning tasks, \jl{i.e.~}entity span detection and relation prediction.
Instead, we train a single network for both tasks simultaneously.
%, since training two different networks for the two tasks separately didn't increase performance in preliminary experiments.

\subsection{Background: Transformers and BERT}
\label{sec:transformers}

\subsubsection{Transformers:}
Transformer~\cite{transformer} networks have been recently proposed for NLP tasks and are fundamentally different from the previously common RNN and CNN architectures.
Compared to RNNs, which maintain a recurrent state, transformers use multi-head self-attention to introduce conditioning on other timesteps.
This enables the parallel computation of all feature vectors in a transformer layer, unlike RNNs, which process the input sequence one time step at a time.
And unlike RNNs, which have to store information useful for handling long-range dependencies in its hidden state, the transformer can access any timestep directly using the self-attention mechanism.

More specifically, transformers consists of several layers of multi-head self-attention with feedforward layers and skip connections. 
Multi-head self-attention is an extension of the standard attention mechanism~\cite{bahdanau}, with two major differences: (1) attention is applied only within the input sequence and (2) multiple attention heads enable one layer to attend to different places in the input sequence.

\af{Let the transformer consist of $L$ layers, each ($l \in \{1,\dots, L\}$) producing $N$ output vectors $\vec x^{l+1}_1, \dots ,\vec x^{l+1}_N$}\dl{, which are then used as inputs in the $l+1$-th transformer layer.} % of the transformer receives the output of the previous layer \af{$\vec x^l_1 \dots, \vec x^l_N $}% with $ i \in \{0, \dots, N\}$ as input.
\dl{The inputs $\vec x^1_1, \dots ,\vec x^1_N$ to the first transformer layer are the embeddings of the input tokens $x_1, \dots, x_T$.}

The attention scores of \af{the \dl{$l$}-th} layer %$l \in \af{\{0,\dots, L\}}$, \af{where} $L$ is the \af{total} number of \af{transformer} layers,
are computed as follows:
{
\begin{align}
a_{l, h, i,j} &= (\vec{x}^l_i W_Q^{(l,h)})^{\top} (\vec{x}^l_j W_K^{(l,h)} )      \label{eq:attn:scores} \enspace, \\
\alpha_{l,h,i,j} &= \frac{e^{a_{h,l,i,j}}}{\sum_{k=1}^N e^{a_{h,l,i,k}}}   \label{eq:attn:alphas} \enspace ,
\end{align}
where $\alpha_{l,h,i,j}$ is the self-attention score for head $h \in  \{1,\dots ,M\}$ in layer $l$ \af{between} position $i$ \af{(corresponding to $\vec x^l_i$)} \dl{and} position $j$ (corresponding to $\vec x^{l}_j$) %(\af{where $i,j \in \{0, \dots, N\}$ and })
 %in layer $l$, 
and is implemented as a softmax of dot products between the input vectors $\vec{x}^l_i$ and $\vec{x}^l_j$, after %\afrm{mapping using the} 
\af{multiplication with the so called} query and key projection matrices for head $h$ of layer $l$ ($W_Q^{(l,h)}$ and $W_K^{(l,h)}$, \af{respectively}). 

Intermediate representation vectors \af{for each input position} are computed as the concatenation of the $M$ heads' summary vectors, each computed as a $\alpha_{l,h,i,j}$-weighted sum of input vectors \af{$\vec{x}^l_1, \dots, \vec{x}^l_N$}, which are first projected using \af{the matrix $W_V^{(l,h)}$}:
\begin{align}
\vec{h}^l_{i} &= [\sum_{j=1}^N \alpha_{l,h,i,j} \cdot \vec{x}^l_j W_V^{(l,h)}]_{h=1..M}    \label{eq:attn:summ} \enspace .
\end{align}
The output
\af{of the $\dl{l}$-th transformer layer \dl{(which is also the input to the $l+1$-th layer)} is then} given by applying a two-layer feedforward network with a ReLU activation function} on $\vec{h}^l_i$, \af{that is:}
\begin{align}
\vec{x}^{l+1}_i &= \max(0, \vec{h}^l_i W^{(l)}_1 + b^{(l)}_1) W^{(l)}_2 + b^{(l)}_2     \label{eq:attn:out}
\enspace .
\end{align}

For more details, that were omitted here, we refer the reader to the work of Vaswani et al.~\cite{transformer} and other excellent resources, like the Illustrated Transformer\footnote{\url{http://jalammar.github.io/illustrated-transformer/}}.

%\dlnote{TODO: explain transformer a little}

\subsubsection{BERT:}
Following previous work on transfer learning form pretrained transformer-based language models~\cite{openaigpt}, Devlin et al. \cite{bert} pretrain transformers on a large collection of unsupervised language data, \af{leading to a model called BERT.}
However, in contrast to a classical, left-to-right language model used by OpenAI-GPT~\cite{openaigpt}, BERT %proposes to
builds on
pretraining a masked language model (MLM).
\af{The} MLM pretraining is done by randomly masking words, \af{i.e.} by \af{randomly} replacing them with \texttt{[MASK]} tokens, feeding the resulting partially masked sequence into the model and training the model to predict the words that have been masked out, given the other words.
This enables BERT's feature vectors to include information both from the preceding tokens as well as the following tokens, whereas the left-to-right LM pretraining of OpenAI-GPT constrained the model to look only at the past.
In addition to the MLM task, BERT is also pre-trained on a sentence pair classification task. Specifically, it is trained to predict whether one sentence follows another in a text.
This pre-training task is useful for downstream tasks such as entailment, which is formulated as classification of sentence pairs, but also for single sentence classification.

BERT for text works as follows. 
Given a sentence (e.g. ``What songs have Nobuo Uematsu produced''), it is first tokenized to (sub)word level using a WordPiece~\cite{wordpiece} vocabulary ($\rightarrow$ [``What'', ``songs'', ``have'', ``no'', ``\#buo'', ``u'', ``\#ema'', ``\#tsu'', ``produced'']). 
More common words are taken as words (``What'', ``songs'', ``have''), while uncommon words are split into subword units (``nobuo'' $\rightarrow$ [``no'', ``\#buo'']).
%, where the hash sign ``\#'' indicates that this token is the next part of a word).
This method significantly reduces vocabulary size and the amount of rare words without dramatically increasing sequence length.
The input sequence is also padded with a \texttt{[CLS]} token at the beginning and a \texttt{[SEP]} token at the end.

The WordPiece token sequence is then embedded into a sequence of vectors. Position\footnote{The use of self-attention requires explicit position indication since this information can not be implicitly inferred, like in RNNs.} (and sequence type\footnote{BERT uses two sequence types: first-sentence and second-sentence, where the latter is only used for sentence-pair inputs and is thus irrelevant for our task.}) embedding vectors are added to the token embeddings.
%to indicate the positions of the tokens.
The resulting embedding vectors are fed through the transformer, which uses several layers of multi-head self-attention and feedfoward layers, as described above. % to identify and integrate relevant context into the next-level representation for each token.
The output vectors for each token can be used for sequence tagging tasks, while the vector associated with the \texttt{[CLS]} token at the beginning of the sequence is assumed to capture relevant information about the input sequence as a whole, since it has been pre-trained for sentence pair classification. 
%The \texttt{[CLS]} representation vector is thus used for sequence classification problems.

\subsection{Entity span prediction}
\label{sec:approach:spans}
%\pt{\st{Given a NL question, we need to identify the span of this input sequence where the subject entity is mentioned.}
In this step, we intend to identify the span of tokens in the input question referring to the subject entity mentioned in it.
Previous works treated this problem as a binary I/O
%\af{[Note: Does this mean binary?]}\dlnote{yes}
sequence tagging problem, and explored the use of BiLSTM, conditional random fields (CRFs), and \af{combined} BiLSTM-CRF taggers.
The sequence tagging model is trained to classify each token in the input sequence as belonging to the entity span (I) or not (O). 
% \pt{correcttext\st{oldunusedtext}}
Instead, we treat span prediction as a classification problem, where we predict the start and end positions of the entity span using two classifier heads.
This approach assumes that only one entity is mentioned in the question and \af{that} its mention is a single contiguous span. 
%the span where it's mentioned is contiguous. 
%\st{This approach is only possible because of the assumptions that there is only a single entity mention in a question and that entity spans are contiguous}.
Formally, the \af{start-position}
%\af{start-and-end-position}classifiers have
\af{classifier has}
the following form:
\begin{eqnarray}
p(i=\mathtt{START} | x_1, \dots, x_N) &= \frac{e^{{\vec{x}^{L+1}_i}^\top \vec{w}_{\mathtt{START}}}}{\sum_{j=1}^{N} e^{{\vec{x}^{L+1}_j}^\top \vec{w}_{\mathtt{START}}}} \enspace , \label{eq:spanpred}
\end{eqnarray}
where $\vec{x}^{L+1}_i$ is the feature vector produced by BERT's topmost ($L$-th) layer for the $i$-th token of the sequence
%timestep 
 and $\vec{w}_{\mathtt{START}}$ is \af{the} parameter vector \af{of} the start position classifier.
End position prediction %follows the same equation
\af{works analogously, applying a}
%but uses a 
different parameter vector, $\vec{w}_{\mathtt{END}}$.

\subsection{Relation prediction}
\label{sec:approach:relations}
Relation prediction can be considered a sequence classification task since the \simplequestions\ task assumes there is only a single relation mentioned in the question.
Thus, for relation prediction, we use BERT in the sequence classification setting where we take the 
feature vector 
\af{$\vec{x}^{L+1}_{\mathsf{CLS}} = \vec{x}^{L+1}_{1}$
}
produced for the \texttt{[CLS]}\footnote{Before using BERT, the input sequence is first tokenized into WordPieces, a \texttt{[CLS]} token at the beginning and a \texttt{[SEP]} token is added at the end.} 
%\af{[Note: Should we just write "first token" and remove all the details in the footnote?]}\dlnote{added some text in background to clarify this}
token \af{at the beginning of the input sequence} and feed it through a softmax output layer to get a distribution over possible relations:
\begin{align}
    p(r=R_i | x_1, \dots, x_N) &= \dfrac{e^{{\vec{x}^{L+1}_{\mathsf{CLS}}}^\top   \vec{w}_{R_i}}}{\sum_{k=1}^{N_R} e^{{\vec{x}^{L+1}_{\mathsf{CLS}}}^\top \vec{w}_{R_k}}}  \enspace ,   \label{eq:relclass}
\end{align}
where 
%$\vec{x}^L_{\mathsf{CLS}}$ is the vector corresponding to the encoding of the \texttt{[CLS]} token (first time step of the transformer encoding) and
$\vec{w}_{R_i}$ is the vector representation of relation $R_i$\footnote{The vector $\vec{w}_{R_i}$ is a trainable \af{parameter} vector, %of parameters, 
unique for relation \af{$R_i$} (and is thus  not \af{presented by} subsymbolic encodings as \af{it is} for example \af{the case} in~\cite{yu2017improved,lukovnikov2017}).
}.

Previous works~\cite{nearlysolved,yin2016,yu2017improved} propose using the question \textit{pattern} instead of the full original question in order to reduce noise and overfitting. They do this by replacing the predicted entity span with a placeholder token.
Doing this would require training a separate model for relation prediction and introduce dependency on entity span prediction. 
In our %\todo{[PT: do we have multiple approach candidates? if not we shouldn't use final bert based ...]} 
BERT-based approach, we chose to train a single %\pt{\st{BERT}}
model to perform both entity span prediction and relation prediction in a single pass. Thus, we do \textit{not} replace the entity span for relation prediction.
We also experimented with training a separate transformer with (1) setting the attention mask for all self-attention heads such that the entity tokens are ignored and (2) replacing the entity mention with a \texttt{[MASK]} token.
However, both methods failed to improve
%in \jl{a}  significant decrease in 
relation classification accuracy in our experiments. %, for reasons that are not yet clear to us.

Training a separate relation classifier network without entity masking yields results equivalent to simply training a single network for both entity span prediction and relation prediction.

\subsection{Logical Form Selection}
\label{sec:approach:final}
%To obtain final results on \simplequestions, we need an additional \jl{to} step select the final logical forms by combining scores from both previous subtasks.
% between the predicted span and labels of Freebase entities.
%\st{For entity linking, we perform entity retrieval using an index lookup, selecting entities whose labels have the highest string similarity with the predicted entity span.}} 
%During retrieval, we also rank entities by the number of outgoing relations the entity has.

%\pt{[NOTE: how do you take 50 candidates when you've already selected entity and even broken ties?]}
%To get the final logical forms, we take the top 50 entity candidates and top 5 relation candidates and consider all combinations between the two sets as possible candidates.
%Then we discard all candidate entity-predicate pairs that do not exist in the given subset of Freebase. Finally, we take the best remaining candidate, ranked by TODO.
To get the final logical forms, we take the top K (\af{where} K=50 in our experiments) entity candidates during entity retrieval and for each, we take the highest-scored relation that is connected to the entity in the knowledge \af{graph}.
We rank the entity-predicate candidate pairs first \af{based on} the string similarity of any of \af{their} entity labels/aliases with the identified span, breaking ties by favouring entity-predicate pairs with \af{predicates with} higher prediction probability \af{under the BERT model}, and the remaining ties are broken by entity in-degree (the number of triples the entity participates in as an object).

\section{Experimental Setup}
% training
We use the small uncased pretrained BERT \nc{model} from a PyTorch implementation of BERT\footnote{\url{https://github.com/huggingface/pytorch-pretrained-BERT}}.
The whole transformer network and original embeddings were finetuned during training.
\af{For training,}
%We used 
the Adam optimizer~\cite{adam} \af{was employed} and \af{we} experimented with different learning rate schedules.
Most of the final results reported use a cosine \dl{annealing} learning rate schedule with a short warmup phase of approximately 5\% of total training updates.
\af{We indicate if reported results rely on a different schedule}.
%Due to long training times and large memory requirements, we performed only a limited hyperparameter search. TODO

We used PyTorch 1.0.1 and trained on single Titan X GPU's.
\dl{The source code is provided at \url{https://github.com/SmartDataAnalytics/semparse-sq}.}

\subsection{Metrics}
To evaluate the entity span prediction model, we compute \jl{the} average\footnote{F1, precision and recall are computed separately for each example based on span overlaps and then averaged across all examples in the dev/test set.} F1 and span accuracy\footnote{Span accuracy is one only for examples where all token memberships are predicted correctly.} on word level.
Since BERT operates on subword-level (WordPiece), \jl{we first need to obtain word-level metrics. To do this, }
%to obtain word-level metrics, 
we first transform the predictive distributions over subword units to distributions over words by summing the probabilities assigned to subword units of a word. 
Then, we take the argmax over the resulting distribution over words.

We also compute F1 on word level over the entire dev/test datasets to compare our numbers to BuboQA~\cite{buboqa}.
%\af{[Note: Here I am confused! Is the difference here, compared to average F1 where you compute the average over the data set?]}
Even though the difference \af{between the} dataset-wide F1 \af{and the} averaged F1 is small, we believe %it is more important to report average F1
\af{the latter is more informative,}
since the contribution of every example is equal and independent of the span lengths.\footnote{The implementation of F1 in BuboQA's evaluation code seems to be computing F1 based on precision and recall computed over the dataset as a whole, thus letting examples with longer spans contribute more towards the final score.} (lower entropy of the predictive categorical distribution)

For relation classification, we report \dl{classification} accuracy.

\subsection{Baseline}
As a baseline, we use a BiLSTM start/end classifier for entity span prediction and a BiLSTM sequence classifier for relation prediction.
The BiLSTM start/end classifier works on word level and uses the same Glove~\cite{glove} embeddings as BuboQA~\cite{buboqa}.
We use the same output layer form as our BERT-based model, where instead of performing a binary I/O tagging of the input sequence, we simply predict the beginning and end positions of the span using a softmax over sequence length (see also Eq.~\ref{eq:spanpred}).
Using this small change significantly improves the performance of our baseline for entity span prediction, as shown in Section~\ref{sec:results}.

For relation classification, we use a different BiLSTM, taking the final state as the question representation vector and using it in a classifier output as in BuboQA~\cite{buboqa} -- comprising of an additional forward layer, a ReLU, a batch normalization layer and a softmax output layer.
We did not replace the entity mentions with an entity placeholder (like~\cite{nearlysolved}), and instead fed the original sequences into the relation classification encoder.

Even though these BiLSTM baselines are quite basic, previous work has shown they can be trained to obtain state-of-the-art results~\cite{buboqa,nearlysolved}.
%\af{[Note: how do that results compare to ours? Or  where they evaluated on different tasks?]}\dlnote{Our baseline is better than or on par with BuboQA. The "nearly solved" paper didn't report the subtask numbers I think. Should I mention this here already?}

Both BiLSTMs were trained using a cosine annealing learning rate schedule as the one used to train our BERT-based model.

\dl{As shown in Section~\ref{sec:results}, our baselines perform better than or on par with equivalent networks used in BuboQA~\cite{buboqa}.%, despite not replacing entity spans for relation classification.
}

%Since we can't use Glove embeddings for WordPiece-tokenized inputs, we use token embedding vectors provided by BERT.
%Thus, the baseline only differs in the parameterization of the actual encoder network.

\subsection{Effect of Limited Training Data}
%In addition, we perform experiments to compare how BERT and \af{the} BiLSTMs perform with less data available for training.
In order to further illustrate the usefulness of fully pretrained models for \simplequestions\ and KGQA, we perform a series of experiments to measure how performance degrades when fewer examples are available for training.
\simplequestions\ is a fairly large dataset containing 75k+ training examples.
With abundant training data available, a randomly initialized model is likely to learn to generalize well, which might make the advantage of starting from a fully pretrained model less pronounced. 
%TODO
The large size of \simplequestions\ makes it possible to study a wider range of \datasparse\ cases than other, smaller datasets.

We run experiments for both BERT and our baseline BiLSTM with different fractions of the original 75k+ training examples retained for training.
Examples are retained such that the number of relations not observed during training is minimized, favouring the removal of examples with most frequently occurring relations. 
We assume that this strategy, compared to random example selection, should not have a big effect on entity span prediction accuracy but should minimize errors in relation prediction due to unseen relation labels, and create more balanced datasets for more informative performance assessment.
We report span accuracy and relation accuracy on the validation set of \simplequestions\ as a function of the fraction of data retained in Table~\ref{tab:spanacc:frac}.
For relation prediction, we only report experiments where the retained examples cover all relations observed in the full training dataset at least once.
%\af{[Note: Where does "seen relations" refer to? Beeing in the testset?]}\dlnote{``Seen relations'' refers to those seen during training, I'll change to ``observed''}

\section{Results and Analysis}
\label{sec:results}
For the two learning tasks, we observe significant improvements from using BERT, as shown in Table~\ref{tab:spanresults} for entity span prediction and Table~\ref{tab:relresults} for relation prediction (see Section~\ref{sec:results:full}).
Section~\ref{sec:results:data} talks about experiments with fewer training data, Section~\ref{sec:results:test} shows component performance on the test set. Final results for the whole simple QA task are discussed in Section~\ref{sec:results:final}. Finally, we conclude with \jl{an} analysis of the attentions in the transformer in Section~\ref{sec:results:attention}.
%\af{That is, we prefer entities with more relation over entities with less.}

\begin{table}[t]
    \centering
    \subfloat[Entity span prediction.
        \label{tab:spanresults}]{
        \centering
        \begin{tabular}{l c c c}
             &  Accuracy & Avg. F1 & F1* \\
             \toprule
    BiLSTM~\cite{buboqa} & -- & -- & 93.1\\
    CRF~\cite{buboqa} & -- & -- & 90.2 \\
    \midrule
    %BiLSTM + Emb  & 92.0 & 96.6 & 96.5 \\
    BiLSTM (ours)        & 93.8 & 97.0 & 97.1 \\
    BERT (ours)          & 95.6 & 97.8 & 97.9 \\
    \bottomrule
        \end{tabular}
        
    }
    \hspace{2em}
    \subfloat[Relation prediction.
        \label{tab:relresults}]{
        \centering
        \begin{tabular}{l c}
             &  Accuracy \\
             \toprule
    BiGRU~\cite{buboqa} & 82.3 \\
    CNN~\cite{buboqa} & 82.8 \\
    \midrule
    BiLSTM (ours) & 82.8 \\
    BERT (ours) &  83.6 \\
    \bottomrule
        \end{tabular}
    }
        \caption{Component performance evaluation results, trained on all available training data, measured on validation set. (a) Entity span prediction performance, measured by span accuracy, average span F1 and dataset-wide F1 (F1*), all on word level.
        (b) Relation prediction performance, measured by accuracy (R@1).}
\end{table}

\begin{table}[b]
    \centering
    \begin{tabular}{l c c c c}
    
    \toprule
         R@N & BiLSTM & BiLSTM & BERT \\
             & \cite{buboqa} & (ours) & (ours) \\
         \midrule
1   & 67.8 & 76.45 & 77.17 & \\
5   & 82.6 & 87.46 & 88.18 & \\
20  & 88.7 & 91.47 & 92.13 & \\
50  & 91.0 & 93.07 & 93.71 & \\
150 & -- & 94.88 & 95.40 & \\
\bottomrule
    \end{tabular}
    \caption{Entity recall on validation set.}
    \label{tab:entityrecalls:dev}
\end{table}

\subsection{Full data results}
\label{sec:results:full}
From Table~\ref{tab:spanresults}, we can see that BERT outperforms our BiLSTM baseline by almost 2\% accuracy (evaluated on validation set), although the difference in F1 is smaller.
Compared to BuboQA~\cite{buboqa}, we obtain much higher dataset-wide F1 scores, which we attribute to our start/end prediction rather than I/O tagging used by previous works, including BuboQA.
%\af{[Note: Could we  test this hypothesis? I.e. could  BERT also  perform tagging?]}\dlnote{Yes, bert could do it too. Will try to run.}

The improvement is less pronounced in relation classification accuracies (see Table~\ref{tab:relresults}), where our baseline BiLSTM achieves the same results as those reported by BuboQA~\cite{buboqa} for \af{a} CNN.
Our BERT-based classifier beats our BiLSTM by almost 1\% accuracy.

Table~\ref{tab:entityrecalls:dev} shows entity retrieval performance for different numbers of candidates, compared against the numbers reported in~\cite{buboqa}. The recall at 50 is 2.71\% higher.
Please note that we also use entity popularity during retrieval to break ties that occur when multiple retrieved entities have the same name (and thus the same string similarity---the main sorting criterion).

\begin{table*}[t]
    \centering
    \begin{tabular}{l l c c c c c c c c c c}
    \toprule
                           & & 0.03\%      & 0.2\%       & 1\%        & 2.5\%        & 5\%         & 10\%         & 25\%          & 50\%          & 75\%         & 100\% \\
                           & & (22)        & (151)       & (757)      & (1k9)        & (3k8)      & (7k6)       & (18k9)       & (37k9)       & (56k8)      & (75k7) \\
    \midrule
%BiLSTM+BERT Emb            & 25.3        & 57.1        & 67.1      &              & 74.9        & 81.3         & 87.0          & 89.3          & 91.0         & 92.0 \\
\multirow{2}{*}{Entity Span} & BiLSTM                     & 33.1        & 64.5        & 74.0       & 78.1         & 82.5        & 85.5         & 90.1          & 92.0          & 93.4         & 93.8\\
 & BERT                       & 62.5        & 79.1        & 85.4       & 88.9         & 90.8        & 92.4         & 94.2          & 94.9          & 95.5         & 95.6\\
\midrule
\multirow{2}{*}{Relation} & BiLSTM & -- & -- & --                    & 26.5  & 41.0        & 56.3         & 72.4          & 79.0          & 81.3         & 82.8\\
 & BERT & -- & -- & --                       & 29.6*  & 48.6        & 67.5         & 76.5          & 80.1          & 82.6         & 83.6\\
\bottomrule
    \end{tabular}
    \caption{Entity span detection accuracies (top half) and relation prediction accuracies (bottom half) as a function of fraction of training data retained. Evaluated on the entire validation set.
    (*) indicates a cosine learning rate schedule with restarts --- in extremely low data scenarios for relation classification, this seems to yield better results than the cosine learning rate \dl{schedule without restarts} that is used everywhere else.}
    \label{tab:spanacc:frac}
\end{table*}

\subsection{Effect of Limited Training Data}
\label{sec:results:data}
From the limited-data experiments for entity span prediction shown in Table~\ref{tab:spanacc:frac} (top part), we can conclude that a pretrained transformer is able to generalize much better with fewer examples. In fact, with only 1\% of the original training data used (757 examples), BERT reaches a best span prediction accuracy of 85.4\% on the validation set, corresponding to an average F1 of 93.2.
In contrast, our BiLSTM baseline achieves only 74.0\% span prediction accuracy on the validation set, corresponding to 88.6 F1.
In an extremely data-starved scenario, with only 0.03\% of the original dataset---corresponding to just 22 training examples---the best validation accuracy we observed \af{for BERT} was 62.5\%, corresponding to 80.9 F1.
In the same setting, we were not able to obtain more than 33.1\% accuracy with our BiLSTM baseline.
Overall, we can clearly see that the degradation in performance with less data is much stronger for our Glove-based BiLSTM baseline.

Limited-data experiments for relation prediction (shown in Table~\ref{tab:spanacc:frac}) (bottom part) reveals that relation classification is more challenging for both our BiLSTM and BERT-based models.
However here too, BERT seems to degrade more gracefully than our Glove+BiLSTM baseline.%, even though with a not as spectacular (??) difference as for entity span prediction.

\begin{table}[b]
    \centering
    \begin{tabular}{l c c c}
    \toprule
         & \multicolumn{2}{c}{Entity Span} & Relation \\
         &  Accuracy & Avg. F1 & Accuracy \\
         \midrule
BiLSTM        & 93.2 & 96.7 & 82.4 \\
BERT          & 95.2 & 97.5 & 83.5 \\
        \bottomrule
    \end{tabular}
    \caption{Component results on test set.}
    \label{tab:results:test}
\end{table}

\subsection{Performance on test set}
\label{sec:results:test}
After identifying good hyperparameters for both our BiLSTM baseline and our BERT-based model using the validation set, we 
%, using only the training data for training
%\af{[Note:  What does this mean? Without looking at the validation set for early stopping etc?]}
%\dlnote{there was no early stopping with cosine schedule, number of epochs was predetermined}
%\af{[Note: Why did you retrained? Did you trained on the joint set of previous trainng and validation set?]}\dlnote{no, just training, just wanted to say: we first searched on validation set and then took the best we could find for validation and then did eval on test.}
\dl{evaluated our models using the same evaluation metrics on the test set.}
Results for both entity span prediction and relation prediction on the test set are reported in Table~\ref{tab:results:test}.\footnote{
Note that the test set contains ``unsolvable'' entries, where the correct entity span has not been identified in pre-processing.
For these examples, we set the accuracy and F1 to zero.
}
As shown in Table~\ref{tab:results:test}, the test set results are close to the validation set results for both models.

%\FloatBarrier
\begin{table}[t]
    \centering
    \begin{tabular}{l c}
        \toprule
         Approach & Accuracy \\
         \midrule
         MemNN~\cite{bordes2015large} & 61.6 \\
         Attn. LSTM~\cite{golub2016} & 70.9 \\ 
         GRU~\cite{lukovnikov2017} & 71.2 \\
         BuboQA~\cite{buboqa} & 74.9 \\
         BiGRU~\cite{dai2016} & 75.7 \\
         Attn. CNN~\cite{yin2016} & 76.4 \\
         HR-BiLSTM~\cite{yu2017improved} & 77.0 \\
         BiLSTM-CRF~\cite{nearlysolved} & 78.1 \\
         \midrule
         BERT (ours) & 77.3 \\
         \bottomrule
    \end{tabular}
    \caption{Final accuracy for the full prediction task on the test set of \simplequestions. 
    \protect\footnotemark
    %[TODO: note about excluded works]
    }
    \label{tab:results:final}
\end{table}

\footnotetext{\cite{turejojic} is not included in the comparison because neither \cite{buboqa} or \cite{nearlysolved} could reproduce the reported results (86.8\%).}

\subsection{Final results}
\label{sec:results:final}
In Table~\ref{tab:results:final}, we compare our final predictions against previous works on \simplequestions.
With our simple entity linking and logical form selection procedure (see Section~\ref{sec:approach:final}), we achieve 77.3\% accuracy on the test set of \simplequestions, beating all but one of the existing approaches.
We suspect that the final score can be further improved by finding better rules for logical form selection, however that is not the goal of this study.

%\subsection{\dl{Error Analysis}}
%\label{sec:erranal}
%Given the final accuracy of 77.3\% obtained on the test set, 
\af{Investigating} the entity and relation prediction accuracies separately, we find accuracies of 82.7\% for entities and 86.6\% for relations.
Comparing the 86.6\% for relation accuracy after re-ranking (Section~\ref{sec:approach:final}) to the 83.5\% (Table~\ref{tab:results:test}) relation accuracy before the re-ranking confirms that re-ranking has helped to reduce errors.
%We further analyze our final predictions on the test set to better understand the errors produced by the presented approach.
\af{By analyzing} the 22.7\% of test examples that were predicted \af{incorrectly}, \af{it turned out} that \af{in} 35\% of those cases both  \af{a wrong relation and a wrong entity had been predicted, in} 41\% only the entity \af{was} wrong and 24\% had only \af{a wrong relation}. 
Of all the cases where the entity was predicted wrong, \af{in 28.6\% cases this resulted from the correct entity missing in the candidate set}. Entity retrieval errors are also correlated with relation errors: of the cases where the correct entity was not among the retrieved candidates, 71.2\% had a wrongly predicted relation, against 55.7\% for cases where the correct entity was among the candidates.

\begin{figure}[h!]
    \centering
    \subfloat[Before fine-tuning]{
    \includegraphics[scale=0.26]{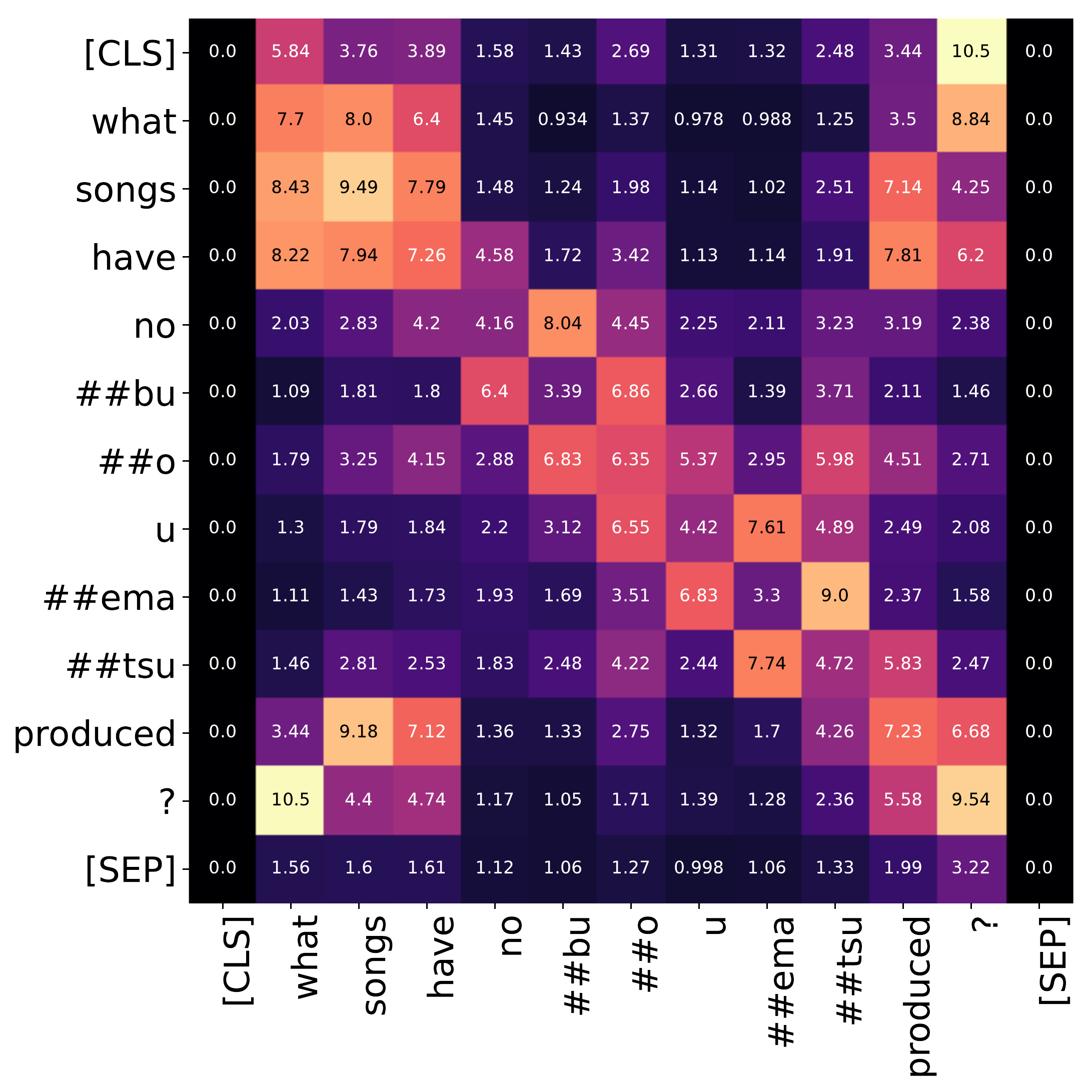}
    }
    \subfloat[After fine-tuning]{
    \includegraphics[scale=0.26]{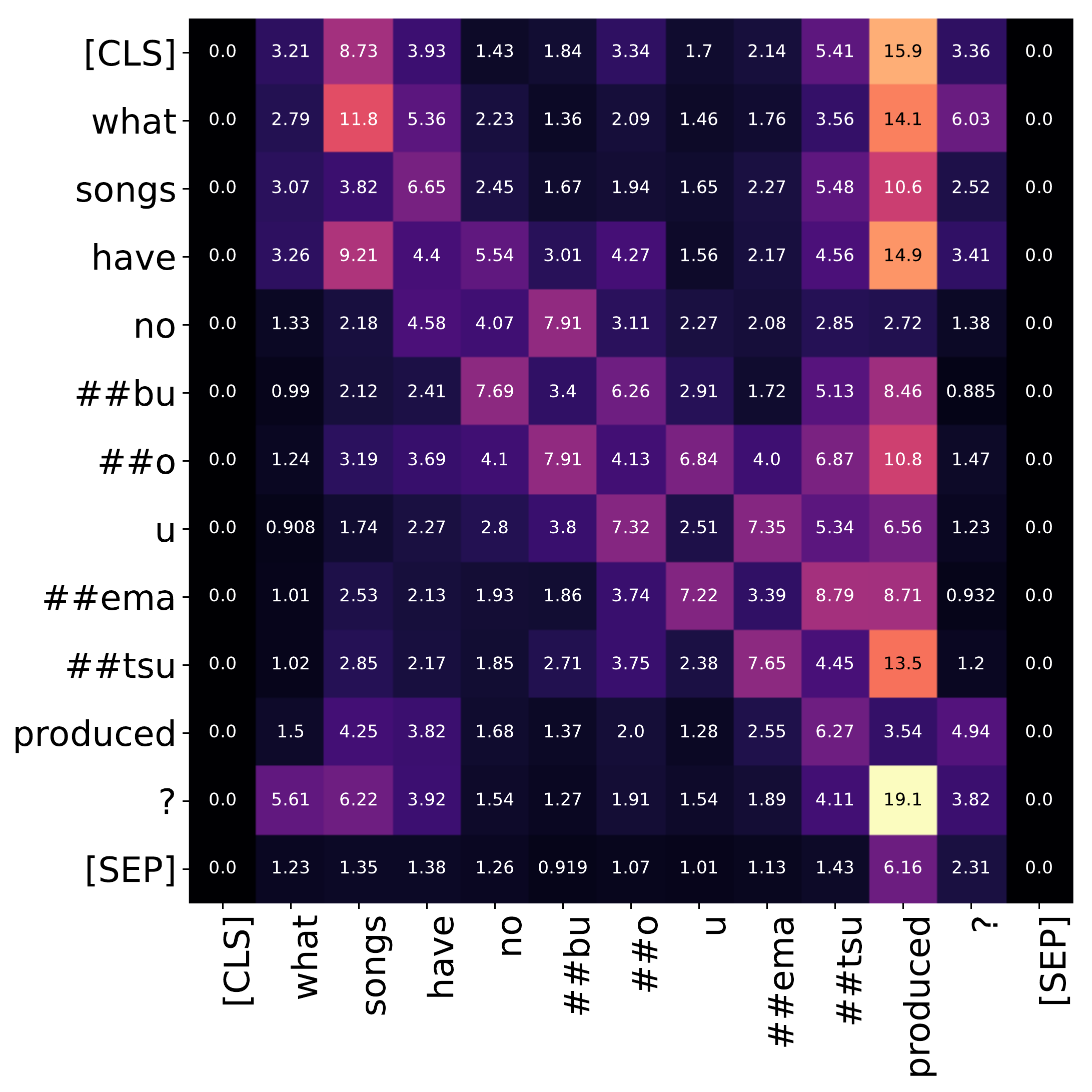}
    }
    \caption{Average attention distribution
    %Attention distributions averaged over all layers and attention heads 
    for the example ``\textit{What songs have \nc{N}obuo \nc{U}ematsu produced?}'', (a) before training on our tasks (vanilla pretrained BERT) \nc{, and} (b) after training on our tasks (finetuned BERT).
    The numbers are scaled \af{to values} between 0 and 100, and are computed \af{by averaging} of the attention distributions over all heads in all layers, \af{and multiplying the average} by 100.
    We set the scores for \texttt{[CLS]} and \texttt{[SEP]} tokens to zero in the plots since they always receive a much higher average attention weight than the actual words from the sentence and thus would dominate the plot.
    }
    \label{fig:att:comp:full}
\end{figure}

\begin{figure}[h!]
    \centering
    \includegraphics[scale=0.26]{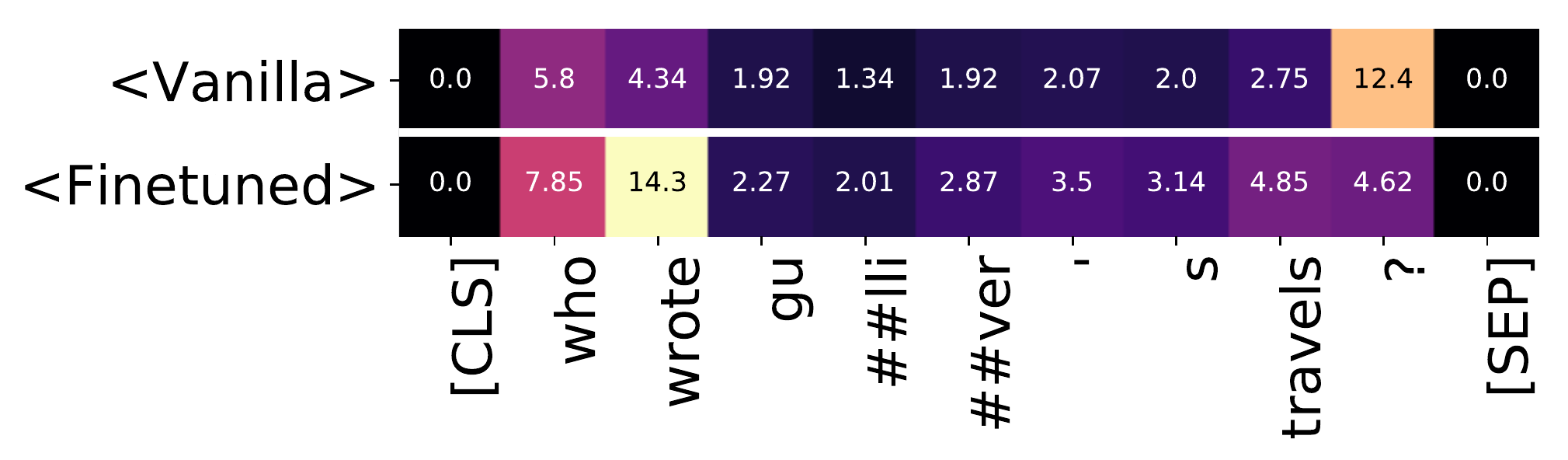}
    \quad
    \includegraphics[scale=0.26]{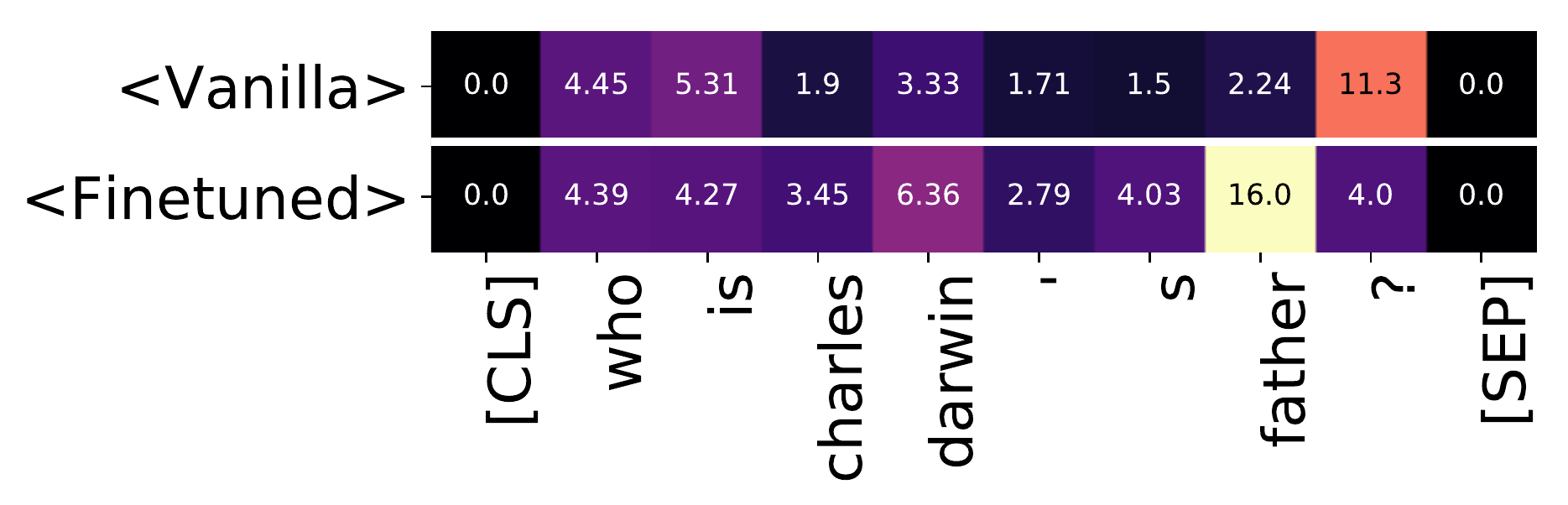}
    \quad
    \includegraphics[scale=0.26]{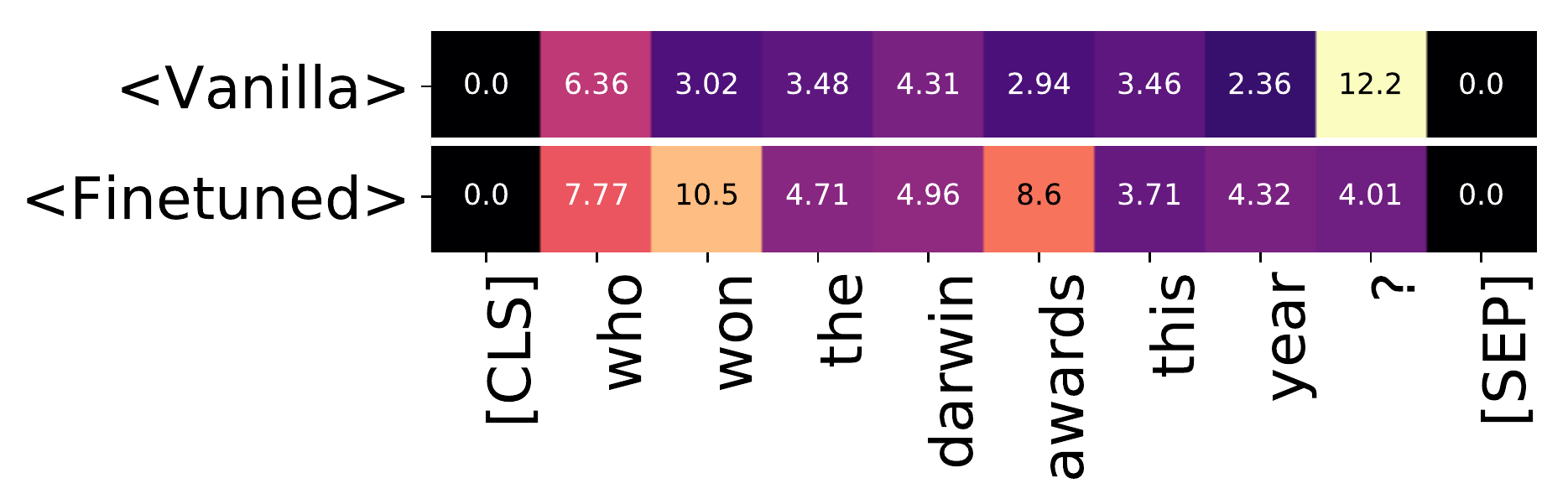}
    \quad
    \includegraphics[scale=0.26]{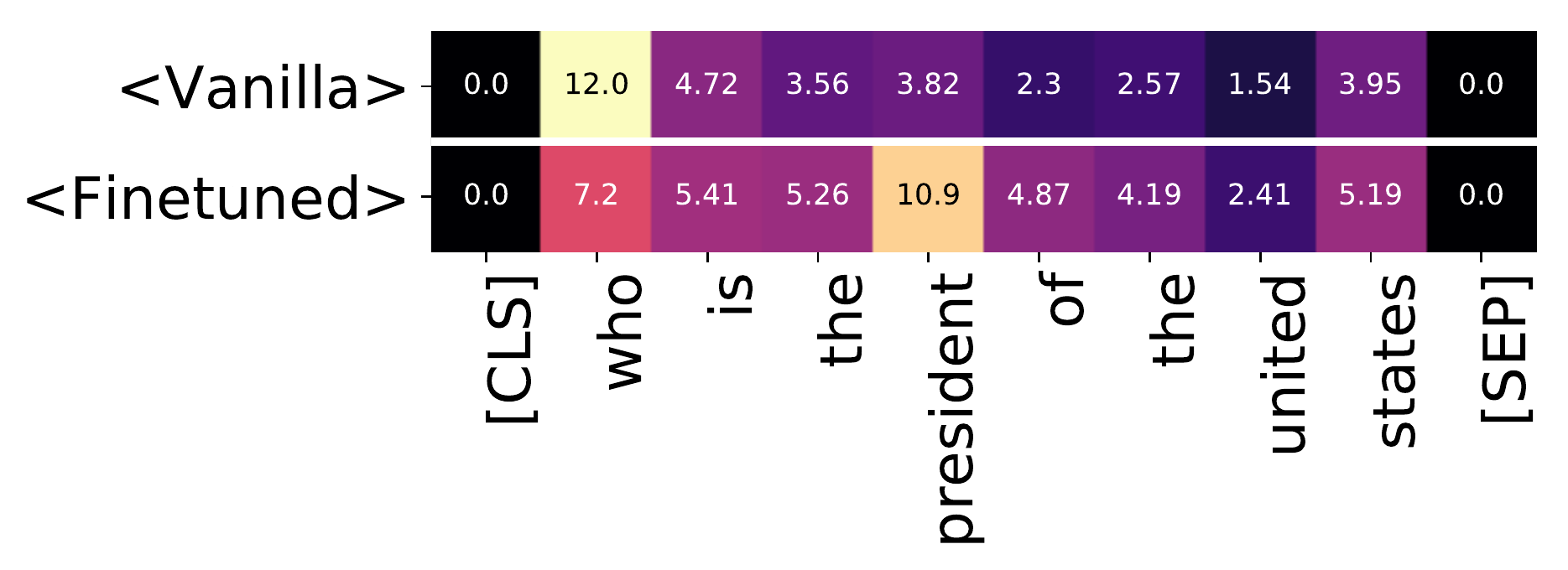}
    \quad
    \includegraphics[scale=0.26]{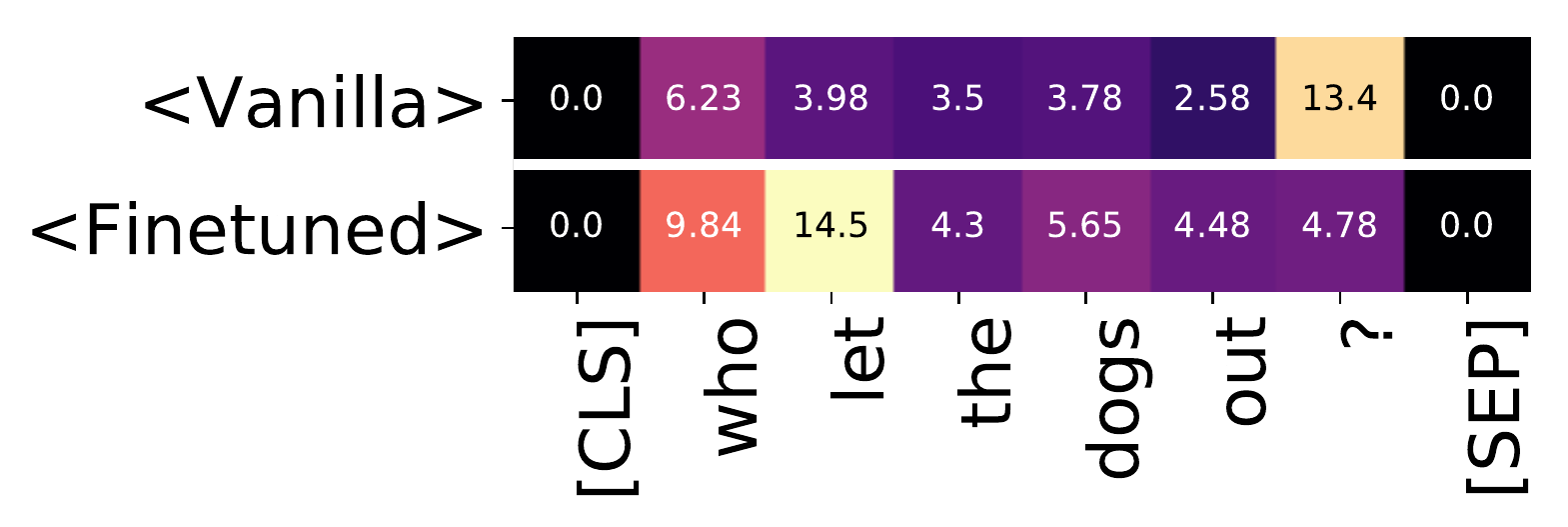}
    \quad
    \includegraphics[scale=0.26]{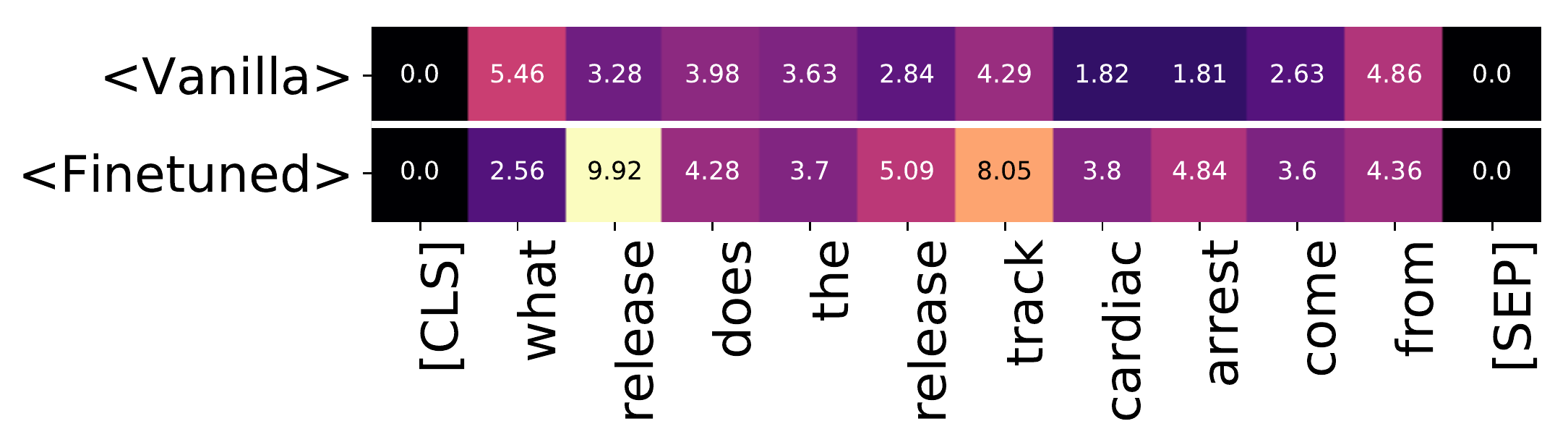}
    \quad
    \includegraphics[scale=0.26]{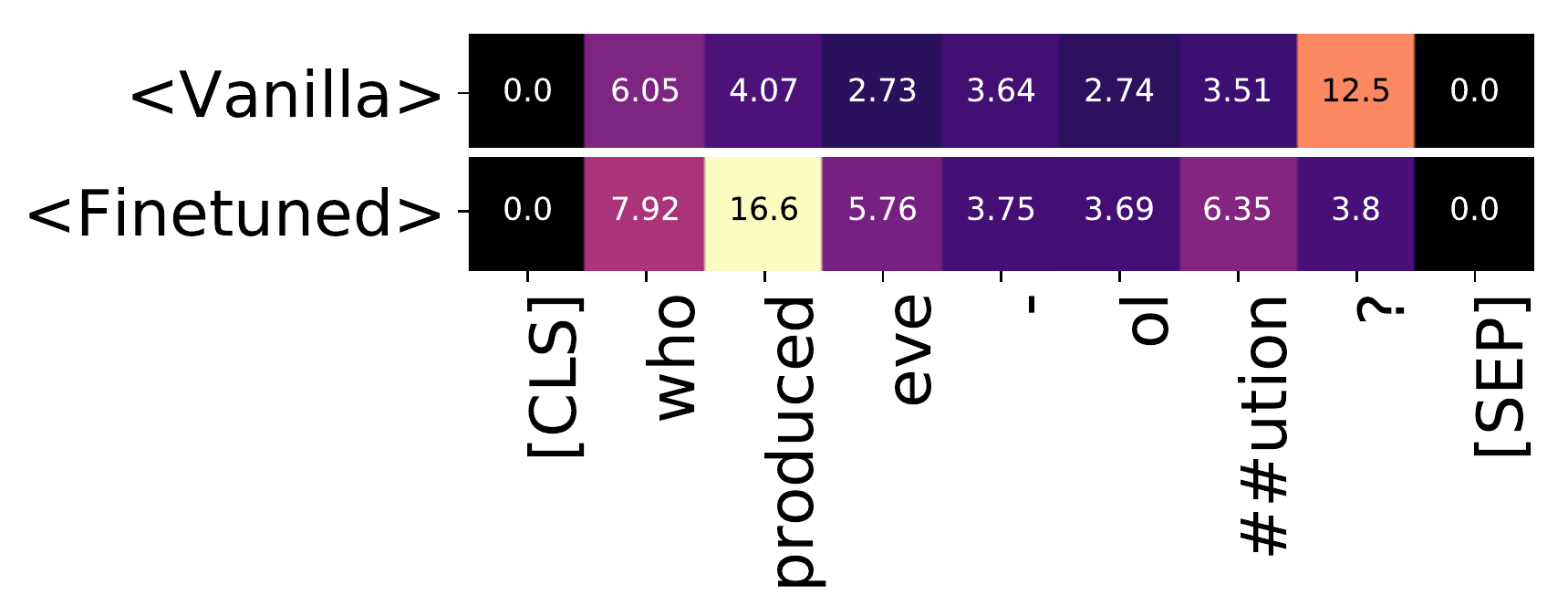}
    \quad
    \includegraphics[scale=0.26]{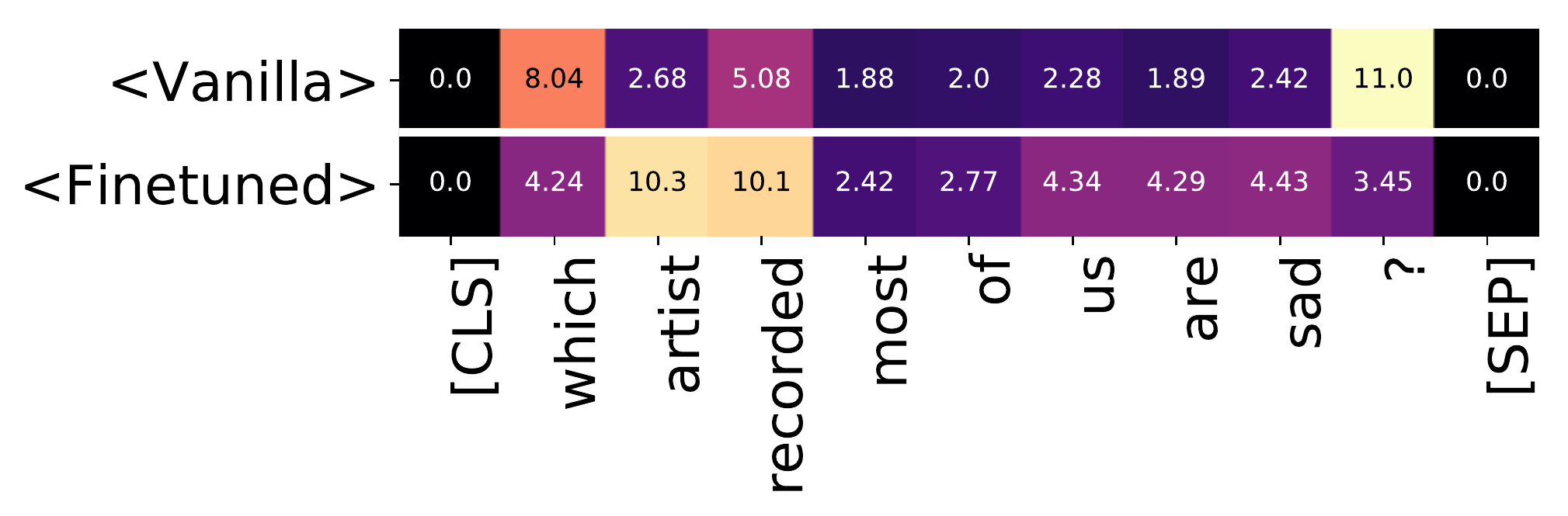}
    \caption{Average attention distributions for the \texttt{[CLS]} token for several examples.
    \texttt{\textless Vanilla\textgreater} is pretrained BERT before finetuning.
    \texttt{\textless Finetuned\textgreater} is BERT finetuned on our tasks.
    The numbers are scaled \af{to values} between 0 and 100, and are computed \af{by averaging} of the attention distributions over all heads in all layers, \af{and multiplying the average} by 100.
    }
    \label{fig:att:comp:cls}
\end{figure}

\subsection{Attention analysis}
\label{sec:results:attention}
One of the advantages of using transformers is the ability to inspect the self-attention weights that the model uses to build its representations.
Even though this does not completely explain the rules the model learned, it is a step towards explainable decision making in deep learning, and a qualitative improvement upon RNNs.
%\dlnote{TODO}
%In comparison, RNNs do not allow for this degree of inspection since the internal dense representations are all that is ex
%higher explainability \dlnote{TODO: explainable decision making motivation}
In an attempt to understand how the model works, before and after fine-tuning, we manually inspected the attention distributions used  by the transformer network internally during the encoding process.

We compute the average of the 144 attention distributions produced by the \af{$M=12$} different attention heads in each of the $L=12$ layers of the \af{employed }BERT network: %internally (Since the BERT network we used has 12 layers, each containing 12 attention heads, which is cumbersome to inspect, we compute the average of the 144 attention distributions.
\begin{equation}
    \label{eq:attanal:avgatt}
    \beta_{i,j} = \frac{\sum_{l=1}^L \sum_{h=1}^M \alpha_{l,h,i,j}}{L \cdot M} \enspace ,
\end{equation}
where $\alpha_{l,h,i,j}$ are the attention probabilities as computed in Eq.~\ref{eq:attn:alphas}, 
%$L$ is the number of layers and $H$ is the number of heads per layer. 
Here, $\beta_{i,j}$'s are the average attention scores; these values are displayed in Figures~\ref{fig:att:comp:full}~and~\ref{fig:att:comp:cls} (multiplied by 100 \af{for scaling}).
\af{More concretly,}
%\dltodo{Add equation?}
we compare this average attention signature of a (vanilla) BERT \af{network} \af{before fine-tuning it }with the attention signature of a BERT model fine-tuned for our tasks (recall that we trained a single model to \af{perform} both border detection and relation classification simultaneously).
By comparing the attentions before and after training on our tasks, we can identify differences in internal behavior of the model that arose during training.

In Figure~\ref{fig:att:comp:full}, the average of all attention distributions is shown for an example question for two versions of the transformer model: pre-trained (vanilla) BERT and BERT fine-tuned for our tasks.
While in general, the average attention distribution roughly follows the same patterns after fine-tuning, we can see that the behavior of the attention mechanism responsible for building the representation of the \texttt{[CLS]} token is significantly different.
We found that, before fine-tuning, the representation building of the \texttt{[CLS]} token generally focuses on punctuation and less strongly, on other words.
After finetuning, \texttt{[CLS]}'s representation is strongly focused on words that characterise the relation conveyed by the sentence.
For example, for the question ``\textit{Who wrote Gulliver's travels?}'' (see Figure~\ref{fig:att:comp:cls}, first example), the attention is shifted towards the word ``\textit{wrote}'', which specifies the authorship relationship of the intended answer with the subject entity (\textit{Gulliver's Travels}) mentioned in the question. 
We provide several other examples of this kind of attention change in Figure~\ref{fig:att:comp:cls}.

This change in internal attention behavior can be explained by the fact that sequence classification for relation prediction is done based on the representation built for the \texttt{[CLS]} token and attending to relation-specifying words more would produce more useful features for the classifier.

%\dlnote{TODO: future work: this stuff could be useful downstream but requires further investigation}
%Focusing the representation of the \texttt{[CLS]} token more on the relation-specifying words thus should improve predictions.

%\dl{[TODO: remove?]} We could not identify attention behavior changes that could be induced by the entity span prediction task.

\section{Related Work}
\label{sec:related}
Bordes et al. \cite{bordes2015large} propose a memory network (MemNN)-based solution to \simplequestions. They use bag-of-words representations for triples and question and train a model that predicts the right triple by minimizing a margin-based ranking loss as defined in the section above. They compute scores between questions and whole triples, including the triple objects. However, triple objects are answers and thus might not be present in the question, which may affect the performance adversely.
The same work introduces the \simplequestions\ dataset, consisting of approximately 100,000 question-answer pairs.

Follow-up works on \simplequestions\ typically predict the subject entity and predicate separately, (unlike \cite{bordes2015large}, \nc{which ranks} whole triples).
\cite{golub2016} explore fully character-level encoding of questions, entities and predicates, and use an attention-based decoder~\cite{bahdanau}. %(see Section~\ref{sec:seq2seq:intro}). 
\cite{lukovnikov2017} explore building question representations on both word- and character-level. \cite{yu2017improved} explore relation detection in-depth and propose a hierarchical word-level and symbol-level residual representation.
Both \cite{dai2016} and \cite{lukovnikov2017} improve upon them by incorporating structural information such as entity type for entity linking. 
\cite{dai2016} and \cite{yin2016} propose an auxiliary BiRNN+CRF sequence labeling model to determine the span of the entity. 
The detected entity span is then used for filtering entity candidates.
Further, \cite{buboqa} investigates different RNN and CNN-based relation detectors and a BiLSTM and CRF-based entity mention detectors. 
%Instead of using neural networks for entity linking, they use simple [TODO]
\cite{nearlysolved} estimates the upper-bound on accuracy for \simplequestions, which is less than 83.4\% due to unresolvable ambiguities (which is caused by the question lacking information to correctly disambiguate entities). 
Both \cite{nearlysolved} and \cite{buboqa} first identify the entity span, similarly to previous works, but disambiguate the entity without using neural networks.
With extensive hyperparameter tuning, relatively basic models and simple heuristics, \cite{nearlysolved} outperformed previous approaches.

\cite{yin2018structvae} proposes a semi-supervised method for semantic parsing based on a structured variational autoencoder that treats logical forms as tree-structured latent variables, and also performs experiments in limited data settings (on \textsc{ATIS} and \textsc{DJANGO}).

\section{Conclusions}
As demonstrated by our experiments, BERT  \af{significantly outperforms} a strong BiLSTM baseline on the learning problems of relation classification and entity span prediction \af{for} simple questions.
Moreover, the pre-trained transformer %degrades better 
\af{shows less performance decrease}
when confronted with fewer training data, as can be seen in our limited-data study, which to the best of our knowledge is the first \af{ever conducted} for \af{the \simplequestions{} data set}.
The final results on the whole \simplequestions{} task are competitive with the current state-of-the-art.

Our comparison of a fully pre-trained transformer to a BiLSTM-based model where only the word embeddings have been pretrained (Glove) might not yield a fair comparison between the two architectures (transformer vs BiLSTM). Further insights could be gained by analyzing the performance of a BiLSTM, which has also been pretrained as a language model (maybe combined with other tasks) in the future.
\af{Here} instead, our aim was to provide evidence that the use of neural networks pre-trained \af{as} language model is beneficial for knowledge graph-based question answering, in particular for  \simplequestions{}, a usecase not included in the original BERT evaluation and, to the best of our knowledge, not yet explored in \af{the} literature.
%\af{[Note: would it be a lot of work to add a pretrained  BiLSTM ot  an untrained BERT to the analysis?]}\dlnote{If untrained BERT means just transformer from scratch, can try and would be nice although hyperparams would be difficult (training is long).}

Even though BERT improves upon our BiLSTM baseline on \simplequestions,
%\af{[Note: When talking about significance we should usually present some sort of significance test!]}
the improvements in the full data scenario might not justify the significantly longer training and inference times and memory requirements.
These practical concerns, however, could be mitigated by practical tweaks and future research. 
%into model compression for large pre-trained networks.
\af{Furthermore,} with fewer data the performance increases \dl{w.r.t. the baseline} become more spectacular, indicating that using pre-trained networks like BERT might be essential for achieving reasonable performance in \af{limited data scenarios. Such scenarios are}
 common for datasets with more complex questions. \af{Therefore,} we believe pretrained networks like BERT can have a bigger impact for %those tasks 
 \af{complex KGQA} (\af{even} when training with all data available).

%\dl{[TODO: incorportate properly, or remove?]}
%PUT BACK IN? \dl{Regarding the diminishing advantage of pretrained BERT compared to baseline with increasing training data size, we hypothesize that non-pretrained models become more likely to learn useful patterns, and/or an upperbound on performance is being reached.}
%\dl{Datasets for KGQA with more complex questions are more challenging to construct, and are significantly smaller than \simplequestions. }
%\dl{While \simplequestions\ is a large dataset, and performance increases while using all data are not that spectacular compared to the baseline, there}
%\af{[Note: Maybe we  should note that this  scenario is the one usually given for more complicated question in KGQA? ]}

% In the future, we plan to provide a better investigation of BERT's decision process and its limitations. We also plan to investigate the robustness of existing methods as well as BERT-based models to typing mistakes and variation in expressing questions (e.g. using synonyms).

%~\cite{TODO-bert-analysis-papers-READ-THEM!!!}.
%\af{[Note: if we find something interesting, we could sell this as "explainable" decision making, throwing one of the recent basswords :D]}\dlnote{do you think section "Attention signature analysis" can use some "explainable decision making"?}
%\dlnote{TODO: more}

\section*{Acknowledgements}

We acknowledge support by the European Union H2020 Framework Project Cleopatra (GA no.~812997). Futhermore, this work has been supported by the Fraunhofer-Cluster of Excellence ``Cognitive Internet Technologies'' (CCIT).
%\dl{We would also like to thank Gaurav Maheshwari and Priyansh Trivedi for their help.}

% \newpage
%\bibliography{acl2018}
\bibliography{bib}
\bibliographystyle{splncs04}

\end{document}